# 2D Visual Place Recognition for Domestic Service Robots at Night

James Mount and Michael Milford, *IEEE Member*

*Abstract*— Domestic service robots such as lawn mowing and vacuum cleaning robots are the most numerous consumer robots in existence today. While early versions employed random exploration, recent systems fielded by most of the major manufacturers have utilized range-based and visual sensors and user-placed beacons to enable robots to map and localize. However, active range and visual sensing solutions have the disadvantages of being intrusive, expensive, or only providing a 1D scan of the environment, while the requirement for beacon placement imposes other practical limitations. In this paper we present a passive and potentially cheap vision-based solution to 2D localization at night that combines easily obtainable day-time maps with low resolution contrast-normalized image matching algorithms, image sequence-based matching in two-dimensions, place match interpolation and recent advances in conventional low light camera technology. In a range of experiments over a domestic lawn and in a lounge room, we demonstrate that the proposed approach enables 2D localization at night, and analyse the effect on performance of varying odometry noise levels, place match interpolation and sequence matching length. Finally we benchmark the new low light camera technology and show how it can enable robust place recognition even in an environment lit only by a moonless sky, raising the tantalizing possibility of being able to apply all conventional vision algorithms, even in the darkest of nights.

## I. INTRODUCTION

With the advent of domestic robots such as robotic lawn mowers and autonomous vacuum cleaners we have only recently seen widespread penetration of robots in the house, even though personal robots have been on the market since the early 1950s [1]. While these tasks may seem to be relatively simple, for a robot they involve a number of challenging problems including navigation. Current navigation solutions [1] combine 1D range sensors or high quality cameras with SLAM [2], or ignore the problem by implementing random movement behaviours instead.

Over the past decade, vision has become the new mainstream sensor for robotic navigation and object classification, due to the rapid increase in camera capabilities and computer processing power. Vision systems provide a variety of cues about the environment, such as motion, colour, and shape, all with a single sensor, and has advantages over other sensors including low cost, small form factor and low power consumption [3], all relevant characteristics in the context of cheap domestic service robots. However, vision-based solutions face multiple challenges including dealing with camera viewpoint and lighting changes or low light, conditions which are common in domestic situations especially if circumstances dictate that robots should operate at night in an unobtrusive manner.

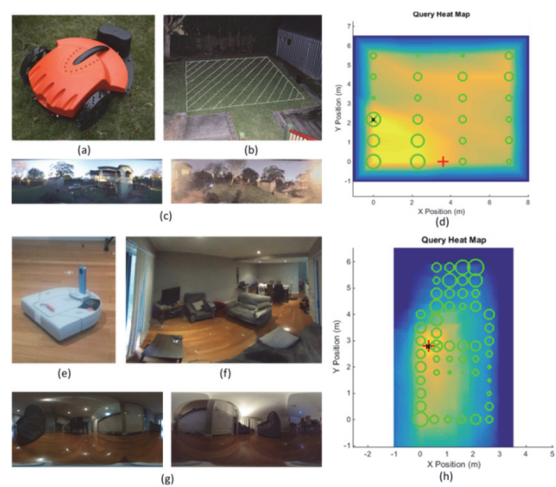

Figure 1: This paper presents a sequence-based place recognition algorithm for two-dimensional localization of outdoor (a-d) and indoor (e-h) service robots under challenging night-time conditions. For the first time, we extend the SeqSLAM algorithm to two-dimensions and combine it with place match interpolation in order to enable place recognition in two-dimensions at night-time.

This paper presents a new 2D localization system for low cost service robots operating both indoors and outdoors, based on low resolution, contrast-enhanced image comparison, sequence-based image comparison in two dimensions, and place match interpolation (Figure 1). The research significantly extends an initial indoors-only proof of concept study with perfect odometry [4], by evaluating the effectiveness of the system in two service robot scenarios; on a domestic home lawn (turf) and inside in a living room environment, analysing the effects of varying odometry noise, the effect of place match interpolation, and introducing new camera technology that enables place recognition in a natural environment lit only by a moonless sky. We also make all the datasets and code freely available online.

The paper proceeds as follows. In Section II we provide a short literature review on autonomous domestic service robots, place recognition approaches for robots, and discuss the nature of the vision invariance problem. In Section III we provide an overview of the approach taken, while Section IV

JM and MM are with the Australian Centre for Robotic Vision and the School of Electrical Engineering and Computer Science at the Queensland University of Technology, Brisbane, Australia, michael.milford@qut.edu.au. This work was supported by an Australian Research Council Future Fellowship FT140101229 to MM.

summarises the experimental setup. Section V presents the results, with discussion in Section VI.

## II. BACKGROUND

Since the 1980s domestic robots have been the subject of intense development, with a focus on ground-based robots such as autonomous lawn mowers and vacuum cleaners. However, it has only been in the past decade that service robots have become widespread on the consumer market, with iRbot, for example, selling an estimated 6 million "Roomba" robots between 2002-2010 [1]. Only simple reactive behaviours such as "edge-following" and "spiral" were implemented on the early systems, while more recent robots implement more sophisticated technologies, including navigation and path planning techniques [1]. Currently autonomous lawn mowers use a variety of technologies including wireless beacons, below surface boundary wires that emit electromagnetic signals, GPS, laser scanners, and even radio technologies [5]. Modern robotic vacuum cleaners today utilize IR, 2D and/or ceiling facing cameras to help navigate and localize within an environment. These inclusions of more advanced technologies have further increased the marketability of such devices by improving task efficiency and lowering operational times and energy consumption.

These improvements have come as a result of the robotic systems now including IR, 2D laser scanners, and/or cameras to help map and navigate an area [1], in conjunction with Simultaneous Localization And Mapping (SLAM) algorithms. SLAM is the process of learning an unknown environment while simultaneously localizing a robot's position. A multitude of mature vision-based SLAM systems are available including MonoSLAM [2], FrameSLAM [6], V-GPS [7], Mini-SLAM [8] and several others [9-16]. However aspects of the problem remain unsolved, such as robustly performing place recognition and loop closure in the face of varying illumination.

Varying illumination and poor lighting are particularly relevant problems for domestic service robots. Firstly, typical homes undergo significant and often unpredictable lighting changes, due not only to day-night cycles and weather variation but also the unpredictable nature of humans modifying the environment, such as by turning lights on and off [17]. In addition, service robots are often required to work in environments at night-time when they are unoccupied by humans. In domestic settings, it is desirable that these systems work in as unobtrusive a manner as possible, ruling out active emittance of light or clicking sonars. The research presented in this paper attempts to address this challenge of developing a cheap vision-based passive solution to 2D localization for indoor and outdoor domestic service robots.

## III. APPROACH

In this section we provide a high level overview of the system architecture (Figure 2). Our approach in this work is based on the assumption that a domestic service robot would occasionally be run during the day in good lighting and with a source of motion information (from either wheel encoders or visual odometry), enabling the robot to gather a reference map of day-time images against which night-time localization can be performed. This approach is a reasonable one, as it is likely that current domestic robots that utilize camera-based systems are fully capable of generating a day-time map, such as the Dyson 360 Eye robot vacuum cleaner.

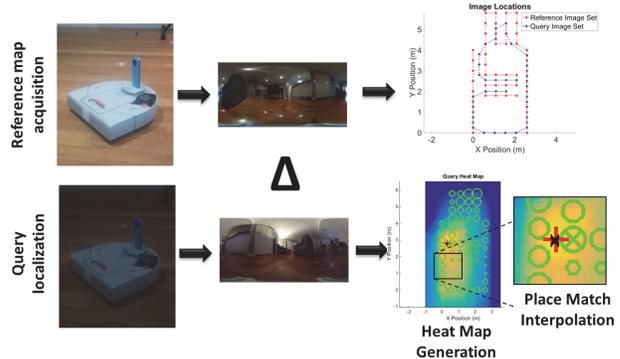

Figure 2: System architecture overview diagram. Reference images acquired in good lighting conditions are mapped to a co-ordinate frame. During night-time operation, a query image is compared with all day-time reference images to generate a heat map, which is interpolated to find the best place match.

### A. Image Set Acquisition

The first step in the process is gathering a reference map of the environment during the day-time, consisting of a topological map and associated camera images at each of the map nodes. We designed a path through the environment with labelled markers for the purpose of ground truthing, and followed this path with a camera, taking images at equally spaced intervals (Figure 3), or by taking a video while walking at a constant velocity. A second set of images was also acquired along a different, only partially overlapping path through the environment, which served as our query / test dataset. All acquired images were also manually mapped to a set of co-ordinates for the purpose of later analysis.

### B. Image Set Preparation

Images were pre-processed using histogram equalization, cropping slightly to remove unwanted features (i.e. the robot or camera mount) and resolution reduced. Finally, patch normalization was performed to reduce the effects of local variations in illumination, such as patches of sunlight on the floor which disappear at night. The patch normalized pixel intensities, $I'$, are given by:

$$I'_{xy} = \frac{I_{xy} - \mu_{xy}}{\sigma_{xy}} \qquad (1)$$

where $\mu_{xy}$ and $\sigma_{xy}$ are the mean and standard deviation of pixel values in a patch of size $P_{size}$ surrounding $(x, y)$.

### C. Image Set Comparison and Score Matrix

Images from each query/test dataset were compared to all images in the reference datasets using a rotation-invariant matching process. Each query image was compared using a sum of absolute differences to every image stored in the reference map, at all possible image rotations. The difference score for the $k^{th}$ rotation, $C(k)$, is given by:

$$C(k) = \frac{1}{h \times w} \sum_{i=1}^{h} \sum_{j=1}^{w} |QS(i,j)_k - RS(i,j)| \Big\|_{k=1}^{k=n} \qquad (2)$$

where $h$ and $w$ are the size of the patch normalized image in the vertical and horizontal directions, respectively. $RS(i,j)$ is the cropped, down-sampled and patch normalized reference set image, $QS(i,j)_k$ is the cropped, down-sampled and patch normalized query set image at the $k^{th}$ rotation, and $n$ is the number of pixel offset rotations. The difference scores between a query image and all the reference images were stored within an image difference matrix.

### D. Heat Map Generation and Best Match Location

To exploit the two-dimensional nature of the environment and enable sequence-matching in two dimensions, a place match heat map was generated for each query image. To generate the heat map, the minimum rotated matching score between the given query image and all reference images was found:

$$MinScores(i) = \min(scores(j)_i) \qquad (3)$$

where $i$ represents the $i^{th}$ reference image compared to the current query image, and where $scores(j)_i$ represents the scores for the current query image against the $i^{th}$ reference set image for all relative image rotations. For visualization purposes, the values within this minimum score matrix were then inverted so that the maximum value "hot spot" corresponded to the best match.

To generate a continous heat map even with irregular reference map image locations, image comparison scores were linearly interpolated across a regular overlaid grid to generate the heat map. Figure 9 shows an example of the resultant regular heat map, showing the response for comparison of a query image against all reference map images. The reference image locations are also plotted with green circles, with the circle size being directly proportional to the matching score for each location. Finally, the interpolated best match position $P$ was found by finding the maximum matching score within the heat map:

$$P = \text{coordinate}\,(\max(InvScores)) \qquad (4)$$

The closest best match reference image was then also determined by finding the closest reference image location to the interpolated location.

### E. Sequence Matching in 2D

Based on the success of sequence-based matching in varying lighting conditions in 1-dimension [16, 18], we developed a two-dimensional sequence matching approach utilizing the heat map. Sequence-based heat maps were generated based not only on the current matching scores, but also on the $n$ previous matching scores, depending on the number of frames used in the sequence.

To generate the sequential heat map, the previous interpolated heat map is taken and translated the same distance as the shift in the query image location from the previous query location and then summed with the current query image heat map. For these experiments we used simulated odometry with varying noise models; for a live robot implementation this data would come directly from either the robot's wheel encoders or a visual odometry system, or both. The best match position and closest reference image match were then found using the same process as for the single frame matching method. All code and datasets are freely available at the following link: https://wiki.qut.edu.au/display/cyphy/Datasets

## IV. EXPERIMENTAL SETUP

This section describes the experimental setup, dataset acquisition and pre-processing, ground truth creation and key parameter values. All processing was performed on a Windows 7 64-Bit machine running Matlab 2014.

### A. Camera Equipment

A Sony A7s with a panoramic lens attachment and Ricoh Theta M15 camera were utilized for the garden and lounge room experiments respectively (Figure 3). The Sony camera is a full-frame DSLR with very large pixel wells, enabling images to be gathered in the garden environment at extremely low lighting conditions in which a standard camera would fail. The Ricoh Theta is a spherical camera that consists of two back-to-back fish eye lenses mounted on a slim body which collect full-field of view imagery.

Although these cameras are not cheap at individual retail prices (in the hundreds to low thousands of dollars range), we note that the subsequent experiments demonstrate that only about 1000 pixels of resolution are required. The A7s sensor comprises 12,000,000 pixels – it is likely a mass production process could generate tiny 1000 pixel sensor arrays (with the same pixel well size) for a tiny fraction of the full frame sensor array price.

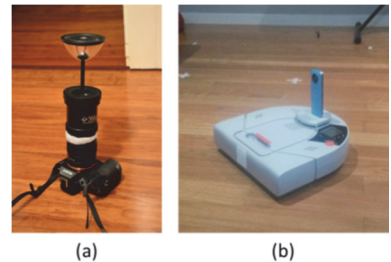

Figure 3: (a) The Sony A7s camera with panoramic lens attachment, used in the Garden experiments. (b) The Ricoh Theta camera mounted on the robot vacuum cleaner used in the living room experiments.

### B. Datasets

Seven datasets were collected in order to run several experiments. Three datasets were taken within the garden environment (Figure 4), two were collected within a living room within a Brisbane townhouse (Figure 5), and two were collected along a creekside path (Figure 6).

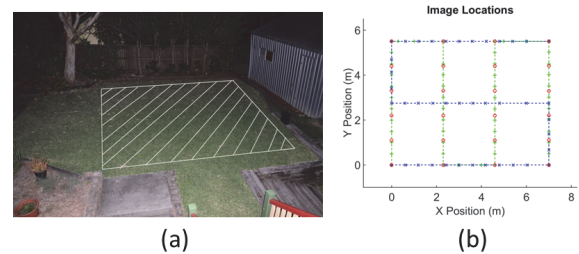

Figure 4: (a) The garden area (illuminated with camera flash) with the white square indicating the area mapped, and (b) the path and image locations for the daytime reference set (red) as well as the night time reference (green) and query sets (blue) in the garden environment.

The first dataset, the daytime reference set, consisted of 24 images in total, over the 7 by 5 metre garden area. The second and third sets, the night time reference and query sets, were taken at night time, with the reference set following the same path as the daytime dataset, while the query set followed an alternative path. There were 53 and 41 images within each of the night sets respectively. The night time images did not necessarily overlap precisely with the reference images, creating a viewpoint invariance problem in addition to the lighting condition-invariance problem.

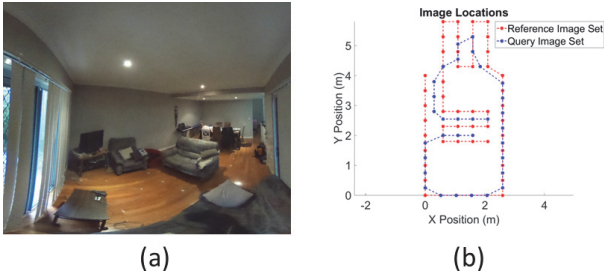

Figure 5: (a) The Living Area environment and (b) the path and image locations for the daytime reference set (red) as well as the night time query set (blue).

The fourth and fifth datasets were taken within a Brisbane townhouse living area. The fourth image set was taken at 52 locations during the day, while the fifth set was taken at 32 query locations in low-light conditions.

The final two datasets were taken along a Creekside pathway during the day-time and in the middle of the night (Figure 6). This environment was chosen because it was mostly completely unlit, with only a moonless sky for illumination. The A7s camera was set to capture video at 10 frames per second at ISO409600 with a lens F stop of 2.0. A summary of all datasets is shown in Table I.

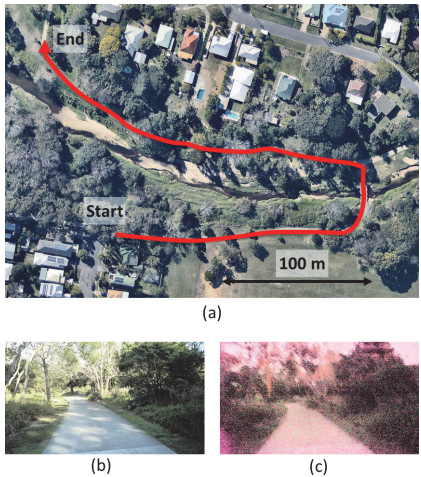

Figure 6: The creek dataset, which involved (a) two 500 metre traverses along a predominantly unlit Creekside path in bushland with only a moonless sky for illumination. Sample (b) day-time frame and (c) night-time frame.

### C. Parameter Values

Parameter values are given in Table I. These parameters were heuristically determined over a range of development datasets and then applied to all the experimental datasets. Table III shows the three Gaussian noise models used when simulating odometry, all noise models had a mean of 0.

Figure 7 shows a visualization of the noise models when applied to a simulated mobile robot.

TABLE I

DATASET SUMMARY

| Name | Size | Frames | Description |
|---|---|---|---|
| Garden Day Reference | 7x5 m | 24 | Daytime reference set, taken in a backyard lawn early morning |
| Garden Night Reference | 7x5 m | 53 | Night time image set, taken in a backyard lawn late at night, and followed the same path as daytime reference set. |
| Garden Night Query | 7x5 m | 41 | Night time image set, taken in a backyard lawn late at night, and followed an alternative path compared to the other image sets. |
| Living Day Reference | 6x3 m | 52 | Reference set was taken during the daytime, with all house lights on, and taken at ground truthed points. |
| Living Night Query | 6x3 m | 32 | Night-time query set was taken during the evening, with two small lamps and oven light on, and were taken at random locations throughout the reference set area. |
| Creek Day Reference | 500 m | 272 | Reference set taken during a normal day with pedestrians and cyclists |
| Creek Night Query | 500 m | 315 | Night-time query set taken with no lighting along the majority of the path, on a moonless night |

TABLE II

PARAMETER LIST

| Parameter | Value | Description |
|---|---|---|
| $R_x, R_y$ | 48,24 | Whole image matching resolution |
| $P_{size}$ | 4 | Patch-normalization radius |
| Interpolated Grid Size | 100,100 | The grid size of the interpolated heat map. |

TABLE III

NOISE MODELS

| Model | Distance Standard Deviation (metres) | Heading Standard Deviation (radians) |
|---|---|---|
| 0 | 0 | 0 |
| 1 | 0.1 | 0.02 |
| 2 | 0.25 | 0.05 |
| 3 | 0.5 | 0.1 |

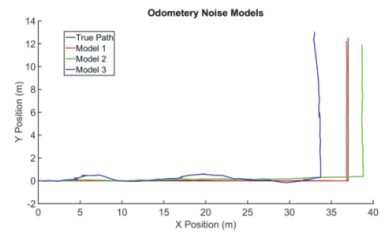

Figure 7: Example short-term trajectories using the three noise models applied when simulating odometry noise. Noise model 1 (red), noise model 2 (green), and noise model 3 (blue).

## V. RESULTS

In this section we present the results of the place recognition experiments. This section is split into 5 parts;

- The Garden Reference place recognition results – which show the image matching results between the garden daytime reference set and the garden night-time reference set, including an analysis of the effect of varying odometry noise and place match interpolation.
- The Garden Query results – which show the image matching results between the garden daytime reference set and the garden night-time query set, involving greater camera viewpoint change, including an analysis of the effect of varying odometry noise and place match interpolation.
- The Living Area results – which show matching performance between the daytime reference set and night-time query set with viewpoint change, including an analysis of the effect of varying odometry noise.
- The Creek place recognition experimental results in near complete darkness
- A comparison of low light camera technology capabilities from 4 years ago to today

There is also a video accompanying the paper illustrating the results.

### A. Garden Reference Results

The results of the Garden reference image matching and place recognition results can be found in the following figures. Figure 8a shows a night time reference set image and its unsuccessful best reference set matched image for single frame matching (Figure 8b), as well as the image it correctly matches to once SeqSLAM is applied (Figure 8c).

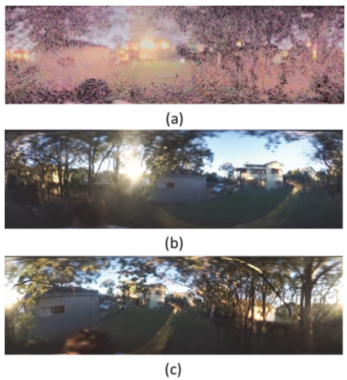

Figure 8: (a) The 12[th] night-time image from the night reference set. Using single image matching, it is incorrectly matched to the 8[th] reference image from the daytime reference set. (c) Using 3 frame 2D SeqSLAM, it is correctly matched to the 6[th] reference image from the daytime reference set.

Figure 9 shows an example of the place match heat maps produced for single frame matching and 7 frame SeqSLAM matching for a particular query image. The reference image locations closest to the query image location (the red-cross) have the maximal matching scores, as indicated by the size of the green circles, but the interpolated place match is more accurate using the sequence-based approach (Figure 9b). Visually it can be seen that the areas within the heat map around the true location of the query image become more prominent in the SeqSLAM heat map compared to the single frame heat map.

Figure 10 shows the distance error, in the form of box plots, using 2D SeqSLAM with sequence lengths from 1 to 10 frames. The distance error is the Euclidian distance between the interpolated location, (the "hotspot"), and the query image's true location. These box plots highlight the improvement of the place recognition algorithm as the matching sequence lengths increase. Figure 11 shows the effect of odometry noise on the 7 frame 2D SeqSLAM implementation, showing a graceful degradation in accuracy as noise increases. Finally, Figure 12 highlights the improvement in place recognition accuracy when place match interpolation is used for the 7 frame SeqSLAM implementation.

Figure 13 shows an image of the Garden environment taken in the same conditions as in which the Night Reference and Query sets were taken, except using the Ricoh Theta camera instead of the low light Sony A7s camera. As can be seen by both the original and the brightened image, the camera is unable to capture any usable information, highlighting the importance of large pixel wells for capturing maximal light.

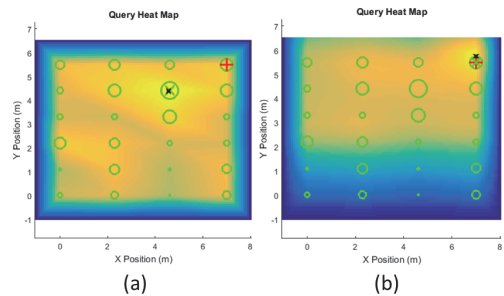

Figure 9: The heat maps for comparison of the 12[th] image in the night-time Garden reference set to all images in the day-time reference set. (a) shows the heat map for single frame matching; while (b) is for 7 frame 2D SeqSLAM. The red-cross shows the ground truth of the query image, while the green cross shows the best matched reference image, and the black cross indicates the best interpolated position (the "hot spot" in the heat map). The green circles are at the coordinates of the reference set image locations, and their size are indicative of how well the current image matches to each reference image.

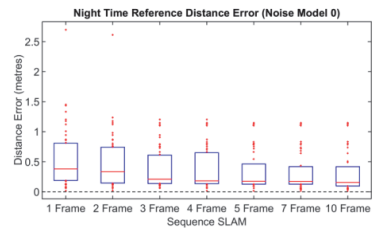

Figure 10: The distance error for place recognition in the night-time Garden reference set using 2D SeqSLAM with variable sequence lengths ranging from 1 to 10 frames.

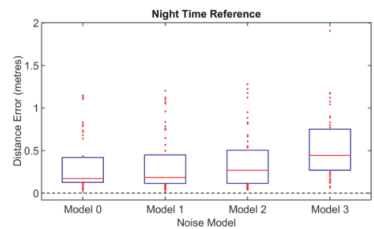

Figure 11: The distance error for the night-time reference set, when compared to the daytime reference set, for 7 frame 2D SeqSLAM with the four varying odometry noise models.

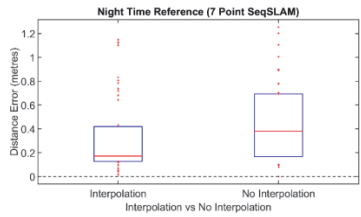

Figure 12: The distance error when interpolation is applied compared to when it is not on the night time reference set for 7 frame 2D SeqSLAM.

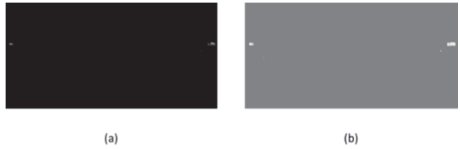

Figure 13: (a) The Ricoh Theta image taken in the Garden Environment at night. (b) The same image as shown in (a) except brightened to highlight that there is no information stored within the image.

### B. Garden Query Results

The results of the Garden query image matching and place recognition results (with more novel viewpoints in the query set) were similar to that of the Garden reference results, and can be found in the following figures. Figure 14 shows an incorrectly matched image using single frame matching, which was successfully matched using the 7 frame 2D SeqSLAM technique. Figure 15 shows the corresponding heat maps.

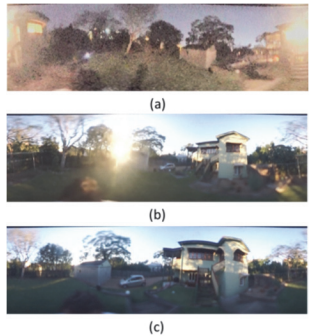

Figure 14: shows the (a) 7th night-time image from the night query set and the (b) incorrectly matched 22nd reference image from the daytime reference set, as well as the (c) correctly matched 12th reference image from the daytime reference set (matched using 7 frame SeqSLAM).

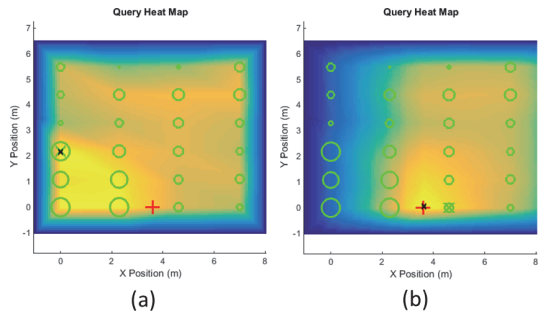

Figure 15: The heat map for the 7th image in the night-time query set. The top heat map is for single frame matching, while the bottom is for 7 frame SeqSLAM.

Figure 16 shows the distance error for the Garden query set across all sequence lengths sets for zero odometry noise, and Figure 17 shows the distance error over the 4 noise models using a 7 frame 2D SeqSLAM implementation. The increased viewpoint novelty does increase the median distance error, but with 10 frame sequence the error is reduced to approximately 0.5 metres. Figure 18 shows the improvement in accuracy caused by place match interpolation.

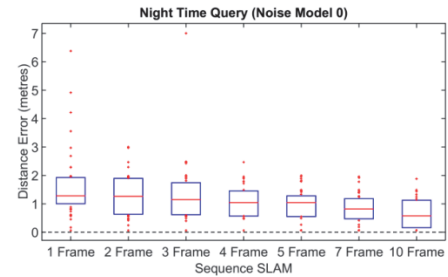

Figure 16: The distance error for the night-time query set, when compared to the daytime reference set, across the 7 sequence lengths with zero odometry noise.

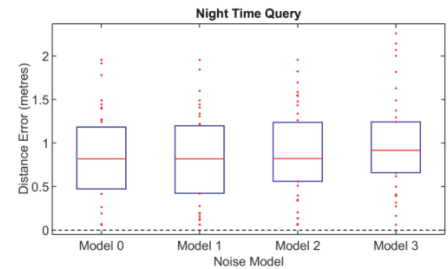

Figure 17: The distance error for the night-time query set, when compared to the daytime reference set, for 7 frame seqSLAM with the four varying odometry noise models.

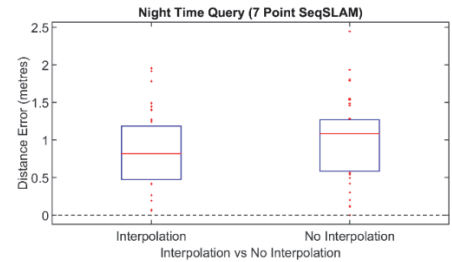

Figure 18: The effect of place match interpolation on the distance error for 7 frame SeqSLAM on the night query dataset.

### C. Living Area Results

We repeated the Garden experiments in the Living Room environment at night. Figure 19 shows the night-time query image incorrectly matching to a daytime image using single image matching, and the correctly matched image when 2D SeqSLAM was applied. Figure 20 shows the corresponding heat maps. Compared to the Gardens dataset, the heat maps have a more localized hot spot due to the increase in the density of the reference dataset.

The distance error for the zero noise model, across the 7 2D SeqSLAM sets can be seen in Figure 21. These results are similar to that of the Garden sets, where the application of SeqSLAM greatly improves the distance error. Figure 22 shows the distance error for 7 frame SeqSLAM across the 4 noise models. Again, similar to the previous Garden sets, increasing levels of noise degrade the median distance error to a maximum of approximately 0.25 metres.

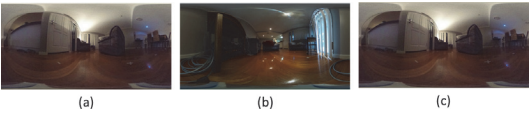

Figure 19: (a) The 28th night-time image from the night query set, incorrectly matched to the (b) 26th reference image from the daytime reference set. (c) Using 4 frame SeqSLAM, it correctly matches to the 41st daytime reference image.

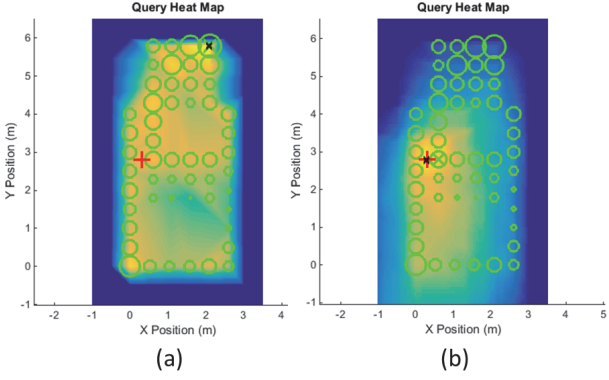

Figure 20: The heat map for the 28th image in the night-time query set for (a) single frame matching and (b) 7 frame SeqSLAM.

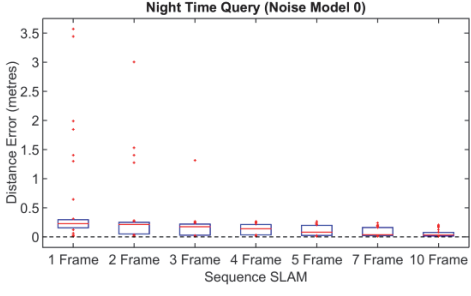

Figure 21: The distance error for the night-time query set, when compared to the daytime reference set, across the 7 sequence SLAM sets with zero odometry noise.

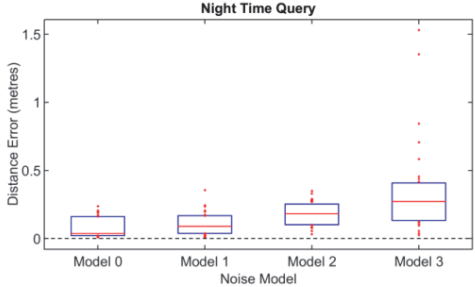

Figure 22: The distance error for the night-time query set, when compared to the daytime reference set, for 7 point seqSLAM with the four varying odometry noise models.

### D. Creek Dataset

Figure 23 shows the frame correspondences found between the two creek path traverses – 81% of places were correctly matched at 100% precision (no false positives, error tolerance of ±3 metres). Figure 24 shows examples of successful place matches. These experimental results demonstrate that current camera technology enables place recognition even in an environment with no lighting beside the moonless sky.

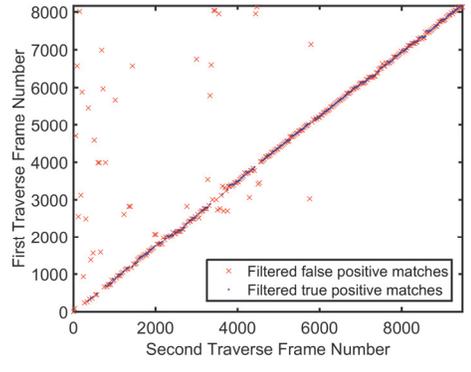

Figure 23: Frame correspondences between the second and first traverses of the creek environment, with blue dots indicating the filtered true positives found using a frame sequence length of 8.

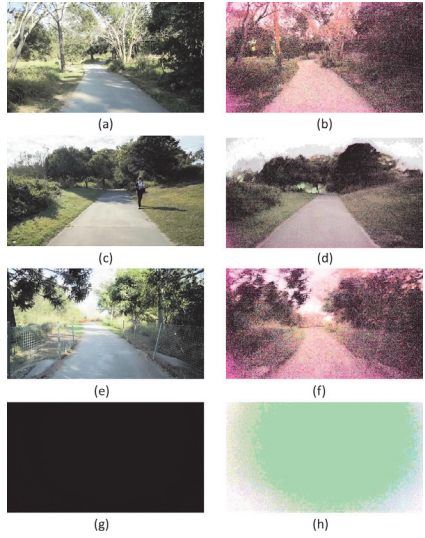

Figure 24: Sample correct frame matches (a-f) found between the day and night traverses of the creek environment, and (g) a comparison long exposure night-mode photo from a current generation smartphone, also (h) shown with brightening.

### E. New Low Light Conventional Camera Technology

Figure 25 shows a comparison between the low light performance of a 4 year old Nikon D5100 DSLR camera and the new Sony A7s DSLR camera used in the Garden and Creek experiments presented here. The new camera technology is capable of achieving acceptable images for localization at high shutter speeds of up to 1/1000s, in conditions where the 4-year-old SLR produces a near black image at only 1/100s shutter speed.

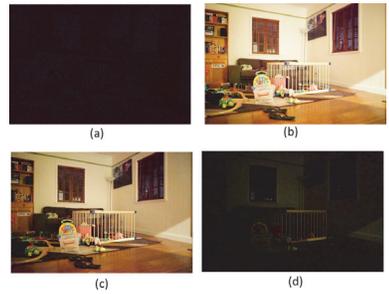

Figure 25: Comparison between a 4 year-old Nikon D5100 DSLR and a current generation Sony A7s in a completely dark room at night. (a) Nikon D5100 at 1/100s exposure, ISO25600. (b) Sony A7s at 1/100s exposure, ISO409600. (c) Sony A7s at 1/500s, ISO409600. (d) Sony A7s at 1/1000s, ISO409600.

*F. Computational Efficiency*

The current algorithms are implemented as unoptimized Matlab code. For the datasets presented here, the primary computational overhead is the image comparison process. When comparing a query image to a dataset of 50 reference images, at a resolution of 48 × 24 pixels at every rotation (48 rotations), just under 3 million pixel comparisons are performed for every query image. A single core CPU can perform approximately 1 billion single byte pixel comparisons per second, while a GPU can do approximately 80 billion per second using optimized C code. Hence the techniques presented here could likely be performed in real-time on a robotic platform when optimized, even on lightweight computation hardware.

## VI. Discussion and Future Work

In this paper we have investigated the potential of low resolution, 2D sequence-based image matching algorithms for performing localization on domestic service robots such as lawn mowing and vacuum cleaning robots in challenging or low light conditions. We assume it is possible to construct an approximate metric map of the environment during the day-time, and use low resolution intensity-normalized image comparison and place match interpolation to match locations experienced at night to the reference day-time locations. While single-matching image performance is relatively poor, using short sequences of a few images significantly improves the average matching accuracy, and is further improved by the place match interpolation process. We have also shown that current camera technology has evolved to the point of enabling place recognition at night in an environment lit only by a moonless sky, opening the possibility in future of robotic vision systems using conventional, passive cameras rather than relying on artificial lighting, thermal sensing or other sensing modalities.

In our current research we are working towards increasing the accuracy of these techniques in order to enable accurate autonomous navigation of cheap indoor and outdoor service robots at night. Estimating image depth, whether using semantic [19] or structure from motion approaches, can lead to increased localization accuracy for whole-image matching approaches. With a method for estimating scene depth, it may also be possible to very sparsely sample the day-time environment and synthetically generate day-time imagery of places not yet visited for the night-time localization system.

Finally, we will investigate how cheaply a robot-specific localization camera can be produced. We have shown that using very low resolution imagery – a few thousand pixels – captured using a sensor with large pixel wells can enable good navigation performance. It may be possible to mass produce custom, much smaller (and hence much cheaper) image sensors with large pixel wells but with an area $1/1000^{th}$ or $1/10000^{th}$ the area of current SLR sensors, vastly reducing their cost and size. Combined with the cheap computational requirements and a motion estimation system, it may be feasible to create a cheap black box place recognition and navigation system that can be deployed on all small robotic platforms, enabling navigation in unlit indoor environments or outdoors on even the darkest of nights.